\documentclass[runningheads]{llncs}
\usepackage{graphicx}
\usepackage{cite}
\usepackage{amsmath,amssymb,amsfonts}
\usepackage{algorithmic}
\usepackage{graphicx}
\usepackage{textcomp}
\usepackage{xcolor}

\usepackage{times}
\usepackage{epsfig}
\usepackage{graphicx,subfigure}
\graphicspath{ {./figs/} }
\usepackage{amsmath}
\usepackage{amssymb}
\usepackage{cite}
\usepackage{indentfirst}

\usepackage[T1]{fontenc}
\usepackage{aecompl}

\def\ie{i.e.}
\usepackage[flushleft]{threeparttable} 
\usepackage{multirow}

\usepackage{bm}

\def\BibTeX{{\rm B\kern-.05em{\sc i\kern-.025em b}\kern-.08em
    T\kern-.1667em\lower.7ex\hbox{E}\kern-.125emX}}

\usepackage{color}
\usepackage{array}
\usepackage{algorithm}               
\usepackage{algorithmic}             

\newcommand{\myparagraph}[1]{\noindent\textbf{#1.}}
\newcommand{\mcr}{\color{red}}

\def\x{\textbf{x}}

\def\M{\mathcal{M}}

\def\ie{i.e.}
\def\eg{e.g.}
\def\ie{i.e.}
\def\eg{e.g.}

\def\0{\textbf{0}}
\def\1{\textbf{1}}

\def\f{\boldsymbol{f}}

\def\x{\boldsymbol{x}}

\def\M{\mathcal{M}}

\begin{document}
%
\title{A Cluster Ensemble Method with Local Mining Metric for Unsuperevised Person Re-identification}
\title{Cluster Ensemble with Local Contrastive Learning for Unsupervised Person Re-identification}
\title{Cluster Ensemble with Contrastive Learning for Unsupervised Person Re-identification}
\title{Contrastive Learning with Cluster Ensemble for Unsupervised Person Re-identification}
\title{Locally Attract, Globally Repel: Contrastive Learning with Cluster Ensemble for Unsupervised Person Re-identification}
\title{Locally Compress, Distantly Repel: Contrastive Learning with Cluster Ensemble for Unsupervised Person Re-identification}
\title{Compress in Ensemble Clustering, Distantly Repel: Contrastive Learning with Cluster Ensemble for Unsupervised Person Re-identification}
\title{Compress Locally, Repel Distantly: Contrastive Learning with Cluster Ensemble for Unsupervised Person Re-identification}
\title{Compress Locally and Temporally: Contrastive Learning via Cluster Ensemble for Unsupervised Person Re-identification}
\title{Compress Locally and Temporally: Contrastive \\ Learning via Multi-Grained Cluster Ensemble for\\ Unsupervised Person Re-identification} 
\title{Contrastive Learning via Multi-Grained Cluster Ensemble for Unsupervised Person Re-identification} 

\title{Contrastive Learning via Multi-Granularity Cluster Ensemble for Unsupervised Person Re-identification} 

\title{Contrastive Learning with Multi-Granularity Cluster Ensemble for Unsupervised Person Re-identification} 

\title{Multi-Granularity Cluster Ensemble based Contrastive Learning for Unsupervised Person Re-identification} 

\title{Multi-Granularity Cluster Ensemble based Hybrid Contrastive Learning for Unsupervised Person Re-identification} 

\title{Hybrid Contrastive Learning for Unsupervised Person Re-identification} 

\title{Hybrid Contrastive Learning with Cluster Ensemble for Unsupervised Person Re-identification} 

%
\titlerunning{Hybrid Contrastive Learning for Unsupervised Person ReID}
%
\author{He Sun\inst{} \and
Mingkun Li\inst{} \and
Chun-Guang Li\inst{}} 
\authorrunning{H. Sun et al.}
%
\institute{School of Artificial Intelligence, Beijing University of Posts and Telecommunications, \\Beijing 100876, P.R. China \\
\email{\{sunhe123, mingkun.li, lichunguang\}@bupt.edu.cn}
}
\maketitle              

\begin{abstract}
Unsupervised person re-identification (ReID) aims to match a query image of a pedestrian to the images in gallery set without supervision labels. 
The most popular approaches to tackle unsupervised person ReID are usually performing a clustering algorithm to yield pseudo labels at first and then exploit the pseudo labels to train a deep neural network. 
However, the pseudo labels are noisy and sensitive to the hyper-parameter(s) in clustering algorithm. 
In this paper, we propose a Hybrid Contrastive Learning (HCL) approach for unsupervised person ReID, which is based on a hybrid between instance-level and cluster-level contrastive loss functions. 
Moreover, we present a Multi-Granularity Clustering Ensemble based Hybrid Contrastive Learning (MGCE-HCL) approach, which adopts a multi-granularity clustering ensemble strategy to mine priority information among the pseudo positive sample pairs and defines a priority-weighted hybrid contrastive loss for better tolerating the noises in the pseudo positive samples. 
%
We conduct extensive experiments on two benchmark datasets Market-1501 and DukeMTMC-reID. Experimental results validate the effectiveness of our proposals. 
%
\end{abstract}
\section{Introduction}
\label{sec:intro}
Person Re-identification (ReID) is a popular and important task in pattern recognition and computer vision, 
aiming to find the images of the same pedestrian in gallery to match the given probe image.
The common approaches are to sort the gallery images according to the similarity between the probe image and the images in the gallery.
Early works are usually based on supervised learning, which trains a deep model with a large amount of labeled data. 
However, the performance of the supervised ReID model will often seriously degenerate when facing the open-world data because the models are usually trained with limited 
data with supervision information. Thus it is crucial to exploit 
the hidden guidance information from the images without supervision. 

In recent years, unsupervised methods for person ReID have attracted a lot of attention. In unsupervised setting, the most popular methods \cite{Fu:ICCV19, GE:ICLR20, zhai:ECCV20, Ge:NIPS20} are based on training a deep neural network with pseudo labels, which are generated by clustering algorithm (\eg, $k$-means, DBSCAN \cite{EsterDBSCAN:AAAI96}). 
For instance, $k$-means is used in \cite{Fu:ICCV19} to generate the pseudo labels for different part of the images and DBSCAN is used in~\cite{GE:ICLR20, zhai:ECCV20, Ge:NIPS20}.

The basic assumption behind the pseudo labels-based unsupervised methods is that 
the samples in the same cluster are more likely with the same class label. Unlike the ground-truth labels, however, the pseudo labels obtained via a clustering algorithm are unavoidably noisy. Thus it is critic to tackle the noises in pseudo labels. For example, in \cite{GE:ICLR20}, a mutual learning strategy via a temporal mean net is leveraged; in \cite{Fu:ICCV19}, a multi-branch network from \cite{wang2018:mgn} is adopted to perform clustering with different part of images. Besides, some works \cite{yu2019:mar, Lin:CVPR20} attempt to exploit 
the neighborhood relationship instead of using traditional clustering methods.

More recently, in \cite{Ge:NIPS20}, contrastive learning is introduced to unsupervised person ReID, in which a hybrid memory bank is used to store all the features and a unified contrastive loss based on the similarity of inputs and all features is adopted to train a deep neural network. While remarkable improvements in performance are reported, all these methods depend upon 
performing clustering method with a delicate hyper-parameter (\eg, the neighborhood ratio parameter $d$ in DBSCAN). Unfortunately, the performance might dramatically degenerate if an improper hyper-parameter is used.





In this paper, we present a simple yet effective contrastive learning-based framework for unsupervised person ReID, in which the noisy pseudo labels are used to define a hybrid contrastive loss---which aims to ``attract'' the pseudo positive samples in the current cluster and at the meantime ``dispel'' all the remaining samples (\ie, the pseudo negative samples) with respect to the current cluster. 
Moreover, we introduce a cluster ensemble strategy to generate multi-granularity clustering information---which is encoded into priority weights, and adopt the priority weights to define a weighted 
hybrid contrastive loss. 
The cluster ensemble strategy aims to alleviate the sensitivity of using a single hyper-parameter in clustering algorithm by using a range of the hyper-parameter to perform clustering ensemble 
instead; whereas the priority-weighting mechanism in the contrastive loss aims to better tolerate the noises in pseudo labels. 
%

\myparagraph{Paper Contributions} The contributions of the paper can be summarized as follows. 
\begin{itemize}
\item We propose a novel hybrid contrastive paradigm for unsupervised person ReID, which is able to better exploit the noisy pseudo labels. 

\item We adopt a multi-granularity clustering ensemble strategy to depict the confidence of positive samples and hence present a priority-weighted hybrid contrastive loss for better tolerating the noises in pseudo positive samples.

\item We conduct extensive experiments on two benchmark datasets and the experimental results validate the effectiveness of our proposals. 

\end{itemize}

\section{Related works}
\label{sec:related-works}

This section provides a brief review on the relevant work in unsupervised person ReID and contrastive learning.

\myparagraph{Unsupervised Person ReID} 
The prior work in unsupervised person ReID can be grouped 
into two categories: a) Unsupervised Domain Adaptation (UDA) based methods and b) pure Unsupervised Learning (USL) based methods. 
UDA is a transfer learning paradigm where both labeled data in source domain and unlabeled data in target domain are 
required. 
However, UDA needs labeled source data and it works only when the distributions of the data in target domain and the data in source domain are closer.   
On the contrary, the USL methods only need the unlabeled data.  
%
%
%
%
Most recent works in USL for unsupervised reID, \eg, \cite{GE:ICLR20, Lin:AAAI19, Zeng:CVPR20, Ge:NIPS20} use 
pseudo labels to train a deep network, in which the pseudo labels are generated by a clustering algorithm, such as $k$-means, DBSCAN~\cite{EsterDBSCAN:AAAI96} and so on. Unfortunately, the pseudo labels are unavoidably noisy, and the clustering results are very sensitive to the hyper-parameter used in the clustering algorithm. 

\myparagraph{Contrastive Learning}
Contrastive learning is a hot topic in recent years. 
Many contrastive learning methods \cite{Oord2018:CPC}\cite{simclr}\cite{MoCov1}\cite{grill2020:BYOL} are developed to learn the hidden information from image samples themselves by minimizing the similarity between different augmented samples of the inputs.
%
%
In \cite{Oord2018:CPC}, InfoNCE loss is proposed and proved that minimizing the InfoNCE loss is equivalent to maximizing the mutual information loss.  In \cite{simclr} and \cite{MoCov1}, a siamese network based framework and a momentum updating paradigm are developed, respectively. 
%
%
%
More recently, contrastive learning strategy has also been introduced to person ReID task, \eg, \cite{Oord2018:CPC, Ge:NIPS20}. Inspired by the InfoNCE loss \cite{wu2018unsupervised}, a unified contrastive loss for UDA based person ReID is presented in \cite{Ge:NIPS20}. 
%
%
%
%
%
%
Different from the previous work, in this paper, we develop a hybrid contrastive learning based unsupervised person ReID baseline at first, and then we present a novel priority-weighted contrastive loss, which effectively encodes a multi-granularity clustering results. 
 



\section{Contrastive Learning based Unsupervised Person ReID}
\label{sec:CLUB}

This section provides some basics on contrastive learning and then present a simple but effective framework for contrastive learning based unsupervised person ReID. 

\subsection{Instance-level and Cluster-level Contrastive learning: A Revisit}
\label{sec:old-baseline}



According to the way to exploit the (pseudo) supervision information, contrastive learning can be divided into two paradigms: a) instance-level contrastive learning, and b) cluster-level contrastive learning. 
Instance-level contrastive learning depends on sample augmentation. Given an input sample, a set of class-preserving samples are generated and fed into a siamese network. 
In such a paradigm, the sample augmentation is assumed to be class-preserving and thus the augmented samples are treated as \textit{positive samples} and all the remaining 
samples in a batch are considered as \textit{negative samples}. Therefore, instance-level contrastive learning mainly leverages self-supervision information from each sample itself individually, without taking into account of the structure or correlation in samples.  

\begin{figure*}
    \centering
    \subfigure[Instance-level]{\includegraphics[width=0.245\columnwidth]{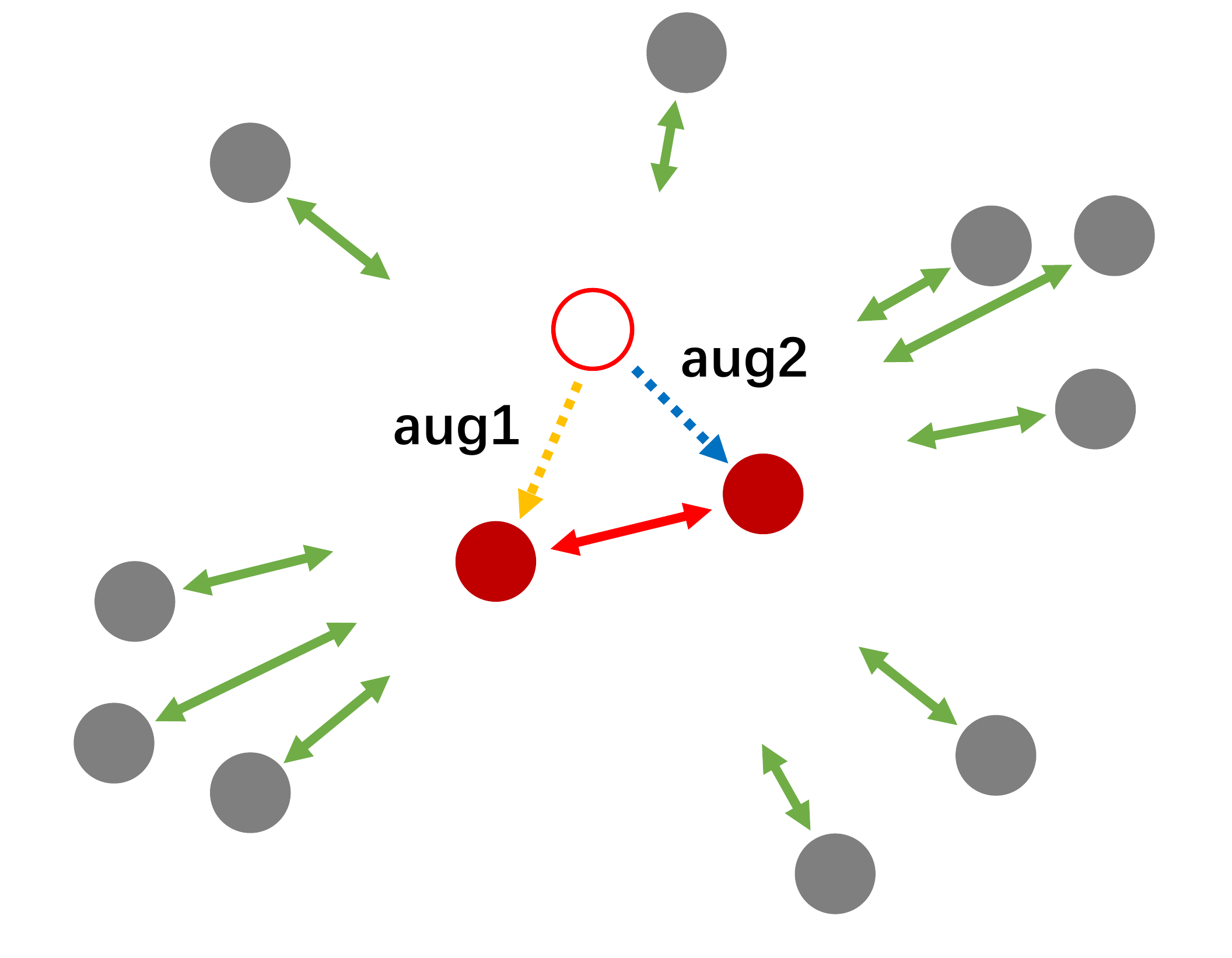}}
    \subfigure[Cluster-level]{\includegraphics[width=0.2450\columnwidth]{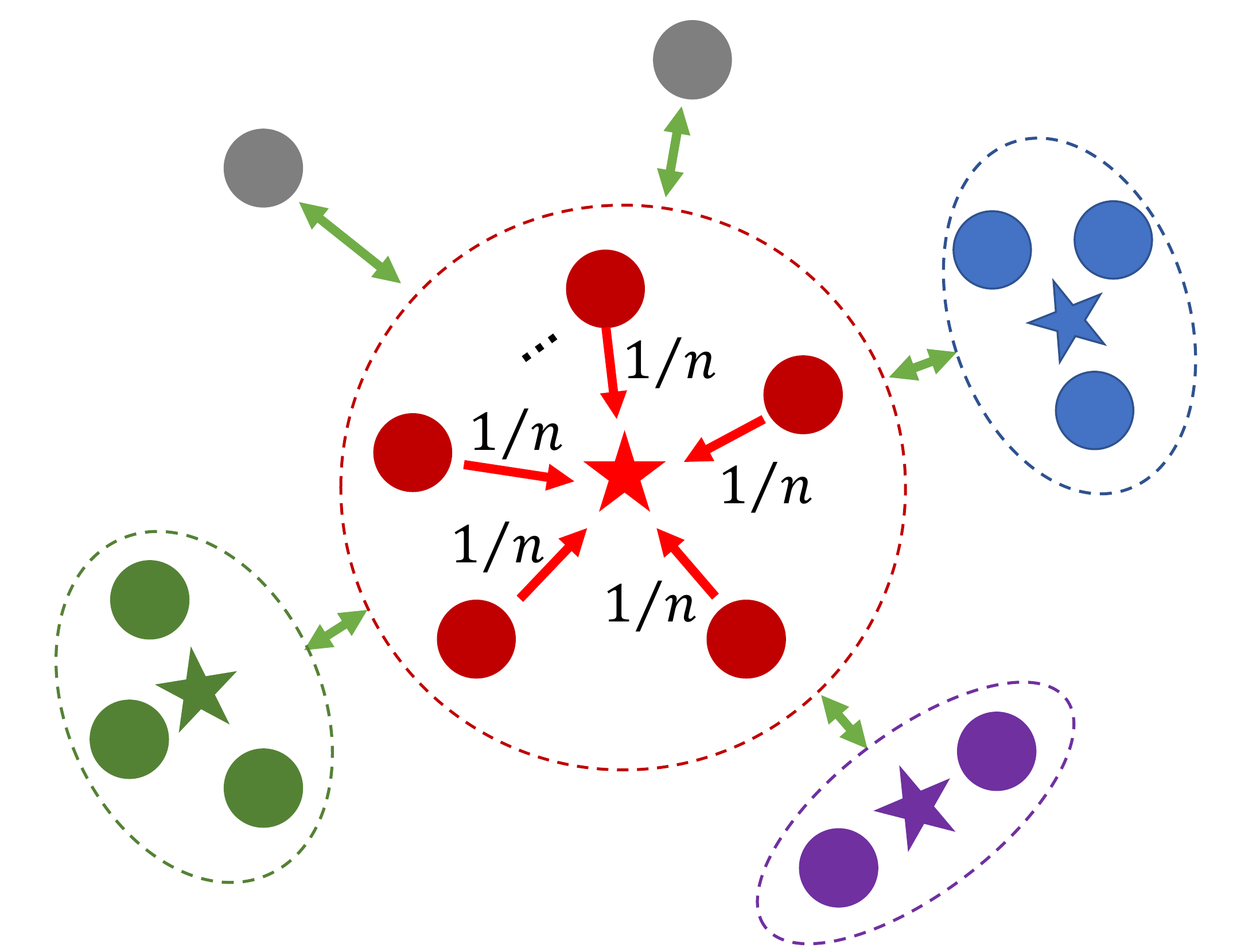}}
    \subfigure[HCL]{\includegraphics[width=0.2450\columnwidth]{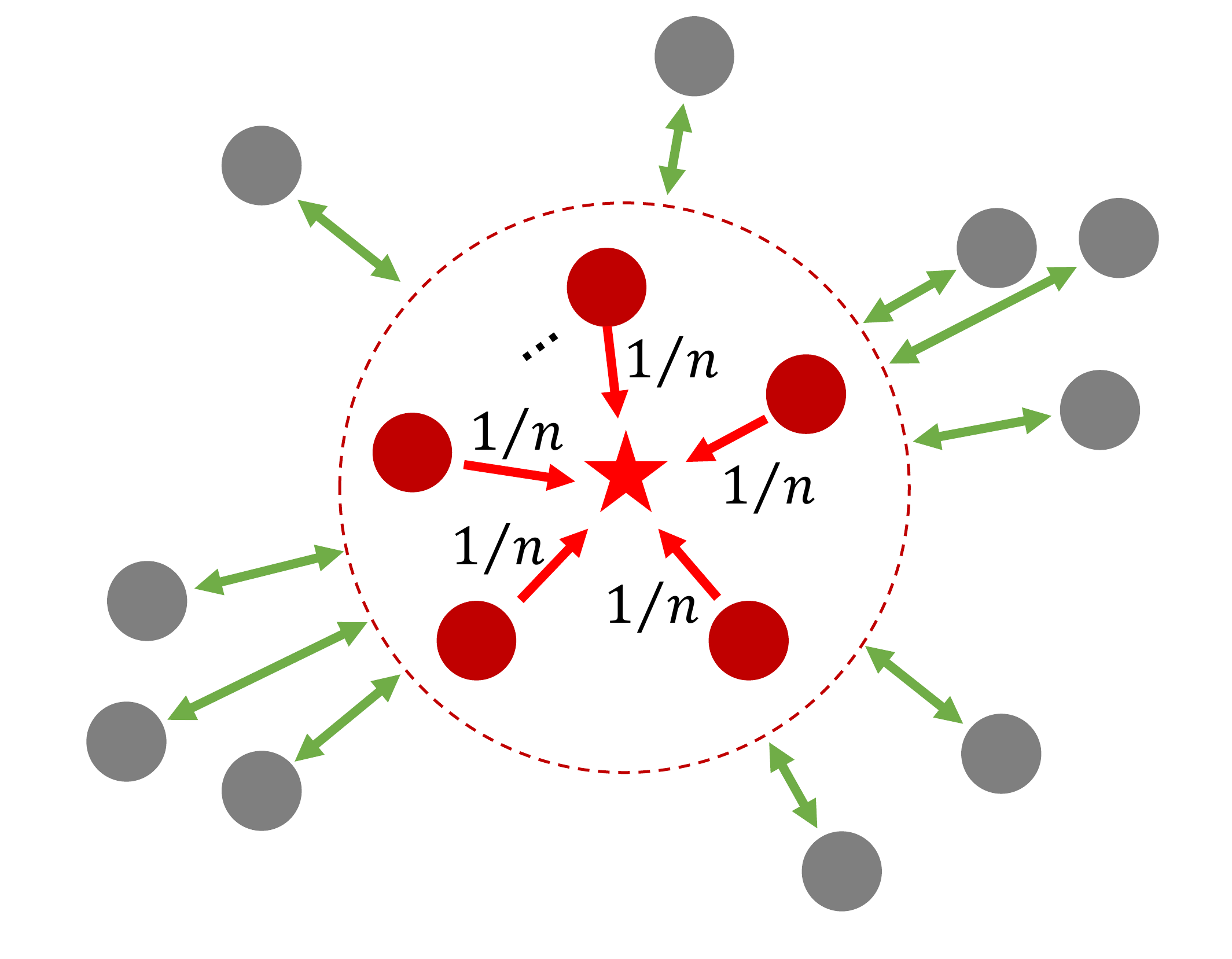}} 
    \subfigure[MGCE-HCL]{\includegraphics[width=0.2450\columnwidth]{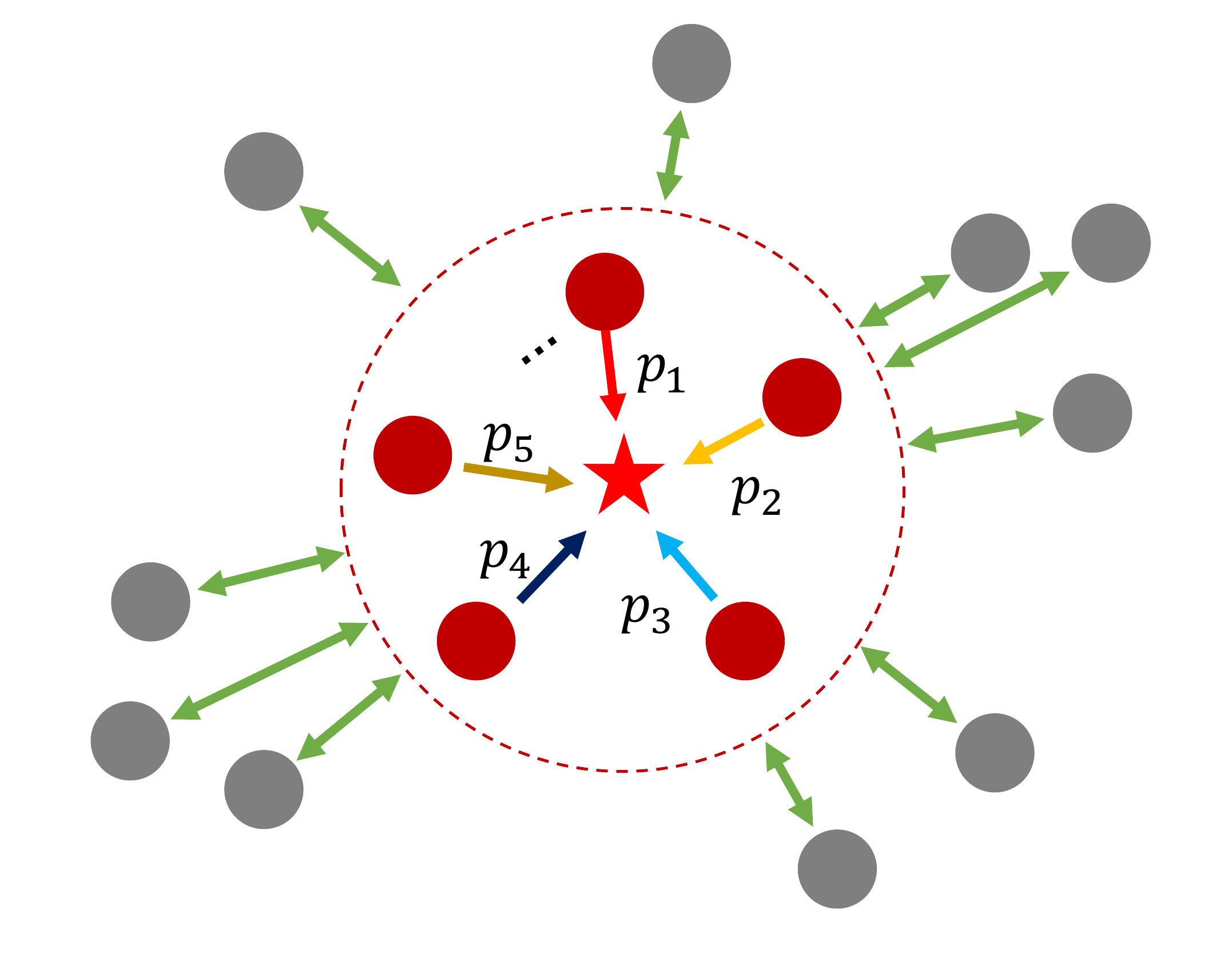}}
    \caption{\textbf{Illustration for 
    different contrastive learning paradigms.} Red arrows in a), b) and c): pulling together; Green arrows: pushing away. (a) For instance-level paradigm, two augmented samples (red points) of original input (red circle) are pulled together and push all others away. (b) For cluster-level paradigm, positive samples (red points) are pushed to the cluster center (red star), and different clusters are mutually exclusive. (c) In our HCL, different from (a) and (b), we consider all negative samples (grey points) individually as individual class. 
    (d) In MGCE-HCL, we further use the priority to weight the similarity between positive samples. 
    }
    \label{fig:CLUB}
\end{figure*}

In cluster-level contrastive learning, cluster information (\ie, pseudo labels), is generated by a clustering algorithm, and the similarity of the feature of the input image and the cluster centers (\ie, the mean vector of each cluster) is used to build an InfoNCE-like loss as follows: 
\begin{align}
\label{eq:contrastive loss}
\mathcal{L}_{con}=-\frac{1}{|\mathcal{B}|}\sum_{i=1}^{|\mathcal{B}|}{\log\frac{\exp(\left\langle\ \bm{f}_{x_i}, \bm{\mu}_+\right\rangle/\tau)}{\sum_{j=1}^{C}\exp(\left\langle\ \bm{f}_{x_i}, \bm{\mu}_j\right\rangle/\tau))}},
\end{align}
where $\tau >0$ is a temperature constant\footnote{By default, we set $\tau=0.05$.}, $\mathcal{B}$ denotes a mini-batch of samples, $|\mathcal{B}|$ denotes the number of the samples in $\mathcal{B}$, $C$ denotes the number of clusters, $\bm{f}_{x_i}$ is the feature representation of an input $\x_i$, and $\f_{M_i}$ denotes the feature from memory bank with index $i$, in which $\bm{f}_{x_i}$ and $\bm{f}_{M_i}$ are defined as
\begin{align}
\label{eq:memory-update}
&\f_{M_i}\leftarrow \gamma \bm{f}_{x_i}+(1-\gamma)\bm{f}_{M_i}, \\
&\f_{M_i}\leftarrow \frac{\bm{f}_{M_i}}{\left\|\bm{f}_{M_i}\right\|_2},
\end{align}
and $\bm{\mu}_{+}$ denotes the mean vector of the samples (\ie, positive sample) of the cluster to which $\x_i$ belongs, and $\bm{\mu}_j$ denotes the mean vector of the samples in the $j$-th cluster.
The reason to use the memory bank $\mathcal{M}$ is 
that the features from memory bank is relative static and thus 
are not only more suitable to perform clustering algorithm for generating pseudo labels but also 
used as a reference for contrastive learning. 
Nevertheless, the dynamic features $\f_{x_i}$ extracted from the backbone are more appropriate for dynamic inputs due to containing 
more information from the random sample augmentation. 


For clarity, we 
illustrate the mechanisms in instance-level contrastive learning and cluster-level contrastive learning in Fig.~\ref{fig:CLUB} (a) and (b). 
Note that both instance-level contrastive learning and cluster-level contrastive learning have shortcomings. Instance-level contrastive learning digs self-supervision information individually for each sample, ignoring the structure or correlation information among samples (\eg, cluster information), which is of vital importance especially for positive samples. 
For cluster-level contrastive learning, while it has been applied to the task such as person ReID, it introduces too much structural information for negative samples, which is usually useless in practice.



\subsection{Hybrid Contrastive Learning (HCL) based Unsupervised Person ReID}
\label{subsec:baseline-CLUB}

To tackle the deficiencies mentioned above, we present a modified contrastive learning paradigm, which is a hybrid between the instance-level paradigm and the cluster-level paradigm, and thus is termed Hybrid Contrastive Learning (HCL).


The HCL framework consists of three components: a) an encoder module for learning convolution feature, b) a memory bank to store the updated features of the whole dataset, and c) a clustering module for generating pseudo labels. We adopt ResNet-50 \cite{He:CVPR2016:resnet} without the full-connection (FC) layer as the encoder module and denote the memory bank as 
$\mathcal{M}=\left\{\bm{f}_{M_i}\right\}_{i=1}^{N}$ which is used to store all the features during the training, where  
$N$ denotes the total number of samples in the dataset. The memory bank is initialized by the normalized features extracted from ResNet-50, which is pre-trained with ImageNet.


In 
training phase, we feed a batch of images, denoted as $\mathcal{B}$, into the backbone and then update the memory bank $\M$ with the new features via Eq.~\eqref{eq:memory-update}, where $\bm{f}_{M_i}$ denotes the feature representation of the sample $\x_i$ in the memory $\mathcal{M}$ and $\f_{\x_i}$ is the convolution feature of the input $\x_i$ extracted by the backbone. 
%
We adopt the DBSCAN algorithm \cite{EsterDBSCAN:AAAI96} with a fixed parameter $d$ to generate the pseudo labels. While the pseudo labels are noisy, there are still rich supervision information 
for contrastive learning. 
%

In this paper, to remedy the deficiencies in instance-level and cluster-level contrastive learning, we propose a hybrid contrastive loss as follows:
\begin{align}
\label{eq:CLUB-loss}
\!\!\!\mathcal{L}_{HCL}=-\frac{1}{|\mathcal{B}|}\!\!\sum_{i=1}^{|\mathcal{B}|}{\log\frac{\exp(\left\langle\ \bm{f}_{x_i}, \bm{\mu}_+\right\rangle/\tau)}{\exp(\left\langle\bm{f}_{x_i}, \bm{\mu}_+\right\rangle/\tau)+\sum_{j \notin \bm{\omega}_+}{\exp(\left\langle\ \bm{f}_{x_i}, \bm{f}_{M_j}\right\rangle/\tau)}}},  
\end{align}
where $\bm \mu_+$ denotes the mean of the positive samples of $\bm{f}_{x_i}$, $|\mathcal{B}|$ denotes the batch size, and 
$j \notin \bm{\omega}_+$ denotes 
to the index set of samples that do not belong to the current cluster $\bm \omega_+$ which corresponds 
to $\bm \mu_+$. For clarity, we illustrate the hybrid contrastive paradigm in Fig.~\ref{fig:CLUB} (c). 
%

\myparagraph{Remark 1} In the modified contrastive loss Eq.~\eqref{eq:CLUB-loss}, we reserve the cluster information for positive samples (\ie, which corresponds to positive cluster) which is able to pull the similar samples together, and at the same time, we treat all the remaining samples---other than the positive samples---as negative samples, rather than using the mean vectors of other (negative) clusters. 
The reasons are two-folds: a) since that the primal goal of the contrastive learning is to pull positive samples together and push all negative samples away, it is not needed to care the cluster information of the negative samples; 
and b) more negative samples are used, more contrasting information can be provided and thus help to avoid obtaining a trivial solution~\cite{2010NCE, Oord2018:CPC}.

\begin{figure*}
    \centering
    \includegraphics[clip=true,trim=0 0 0 0,width=0.96\textwidth]{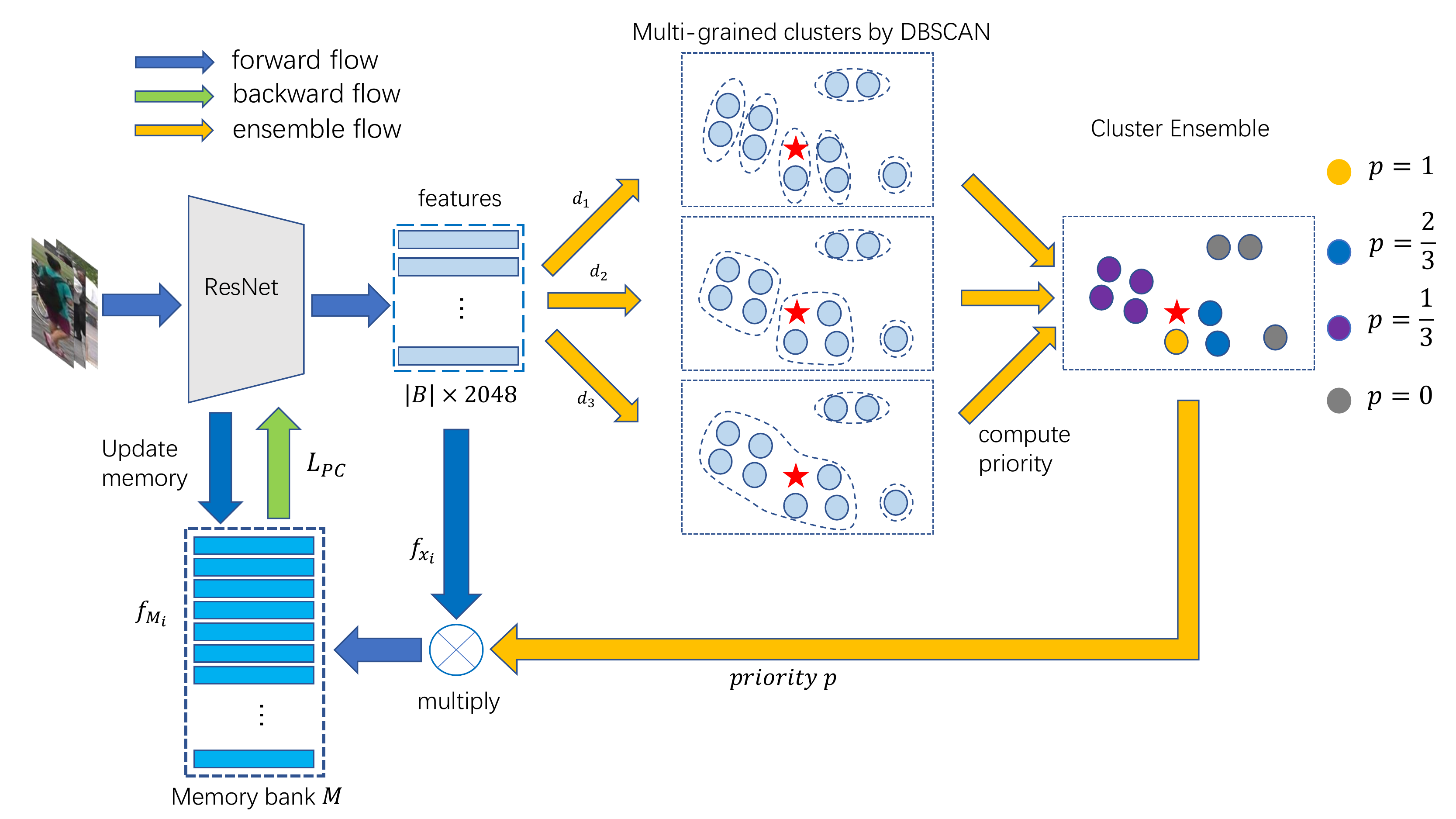}
    \\[-5pt]
    \caption{\textbf{Architecture of MGCE-HCL}. The figure shows the case that three clustering results are 
    assembled. Each batch of images are fed into ResNet50 to obtain the features, 
    and then DBSCAN with 
    parameter $d$ is used to perform clustering with the features in the memory bank. After that, we compute priority matrix with multiple granularity 
    cluster results. In Cluster Ensemble (CE) module, different colors of points denote different value of priority and the {\color{red}red} star denotes the convolution features of the current sample. 
    To compute $\mathcal{L}_{PC}$, we compute the cosine similarity between the input features and the features in memory bank with priority-weighting mechanism.   }
    \label{figure:architecture} 
\end{figure*}

\section{Hybrid Contrastive Learning with Multi-Granularity Cluster Ensemble (MGCE-HCL)}
\label{sec:MGCE-HCL}

This section presents a Hybrid Contrastive Learning framework with Multi-Granularity Clustering Ensemble (MGCE-HCL) for unsupervised person ReID.

Compared to HCL, the key differences of MGCE-HCL are two-folds: a) rather than performing DBSCAN with a single neighborhood parameter $d$ to yield a clustering result, we perform DBSCAN multiple times with parameter $d$ sampled in a range to generate a multi-granularity clustering results and encode the obtained clustering results into priority weights; and b) we introduce the priority weights into the hybrid contrastive loss, which automatically exploits the confidence of the positive sample pairs. 




\myparagraph{Multi-Granularity Clustering Ensemble (MGCE)} To remedy the sensitivity to the hyper-parameter in DBSCAN, we perform  DBSCAN $T$ times, each time using a different neighborhood parameter $d^{(\ell)}$, in which $\{{d^{(\ell)}}\}_{\ell=1}^{T}$ are sampled from a range with an interval $\delta$. 
Let $\bm{c}^{(\ell)}$ be the obtained cluster index of the $\ell$-th clustering with parameter $d^{(\ell)}$, where $\ell=1,\cdots,T$. Accordingly we define an affinity matrix $A^{(\ell)}$, which is calculated as follows: 
\begin{align}
\label{eq:affinity}
\begin{split}
  &A_{i,j}^{(\ell)}=\left \{
\begin{array}{ll}
    1, ~~~{c}_i^{(\ell)}={c}_j^{(\ell)} \\
    0, ~~~{c}_i^{(\ell)}\neq {c}_j^{(\ell)}.
\end{array}
\right. 
\end{split}
\end{align}
Note that $A_{i,j}^{(\ell)}$ is obtained with the neighborhood parameter taking the value $d^{(\ell)}$ and thus we view $A_{i,j}^{(\ell)}$ 
as the affinity under a specific granularity indexed with $d^{(\ell)}$. By taking an average over 
the $T$ results, 
we have a \textbf{priority} weight as follows:
\begin{align}
\label{eq:priority}
p_{i,j} = \frac{1}{T}\sum_{\ell=1}^{T}A_{i,j}^{(\ell)},
\end{align}
where $0 \le p_{i,j} \le 1$ approximately quantifies the confidence of two samples being 
grouped into the same cluster.

\myparagraph{Remark 2} From a geometrical perspective, the priority is 
to describe the neighboring relationship 
between any two samples. This is because that if two samples lie close enough---they are more likely to be grouped into 
the same small 
cluster and thus are certainly to be grouped into larger cluster, 
resulting 
a higher priority value according to Eq.~\eqref{eq:priority}. 
From the 
probabilistic perspective, the priority also measures 
the confidence that the two samples are in the same cluster. 
Briefly, the higher the priority is, the two samples are more likely to be closer and the sample pairs are more credible to be positive samples. On contrary, when the priority of two samples is 
0, it is 
reasonable to consider them as negative samples.

%


\myparagraph{Priority-Weighted Hybrid Contrastive Loss} 
Given the priority weights, we propose a priority-weighted hybrid contrastive loss as follows:
\begin{align}
\label{eq:PC-loss}
\begin{split}
%
\mathcal{L}_{PC}= -\frac{1}{|\mathcal{B}|}\sum_{i=1}^{|\mathcal{B}|}{\log\frac{s_i^{+}}{s_i^{+}+s_i^{-}}},
\end{split}
\end{align}
%
%
in which $s_i^{+}$ and $s_i^{-}$ are defined as the exponential similarity between the input and the positive samples and between the input and the negative samples, respectively, \ie,
\begin{align}
\label{eq:PC-sp}
&s_i^{+}=\exp(\frac{\sum_{j=1}^{N}{p_{i,j}\left\langle \f_{x_i},\f_{M_j}\right\rangle/\tau}}{\sum_{j=1}^{N}{p_{i,j}}}), \\
\label{eq:PC-sn}
&s_i^{-}=\sum_{j=1}^{N}{\mathbb{I}(p_{i,j}=0)\exp(\left\langle \f_{x_i},\f_{M_j}\right\rangle/\tau)},
\end{align}
where $\mathbb{I}(\cdot)$ is an indicator function, $\mathbb{I}(p_{i,j}=0)$ outputs $1$ if $p_{i,j}=0$, and $\left\langle\cdot, \cdot\right\rangle$ denotes the inner product. 
Note that $s_i^+$ is computed by the inner product between the feature of the input image and the features from the memory bank $\mathcal{M}$ and are weighted by the nonzero priority; whereas $s_i^-$ is computed by
the samples whose priority being 
$0$ which are 
considered as negative samples and 
each negative sample pair is treated individually ignoring their cluster information.

For clarity, we provide 
the flowchart of the MGCE-HCL framework in Fig.~\ref{figure:architecture}. 
The input image is shown as the {\mcr red star} in the cluster ensemble module. After obtaining the confidence of all sample pairs, we 
weight each sample pair with the accumulated priority to train the whole 
model.


\myparagraph{Remark 3}
Note that priority-weighted similarity 
defined in Eq.~\eqref{eq:PC-sp} can bring more information for 
positive samples pairs because, the priority is able to describe the density of samples. 
%
%
For the computation of positive scores, we use priority to weight the positive samples of different distance but the cluster-level contrastive loss and HCL only compute the similarity between the input and the mean vector of all positive samples, which is equivalent to give each positive sample the same weight.
For the negative samples, both MGCE-HCL and HCL consider each negative sample individually without using cluster information. 
For clarity, the difference between HCL and MGCE-HCL is illustrated 
in Fig.~\ref{fig:CLUB} (d).


%


\section{Experiments}
\subsection{Datasets and Evaluation Metrics}

\myparagraph{Datasets} 
We evaluate our method with two benchmark datasets: Market-1501\cite{Zheng:ICCV15} and DukeMTMC-reID\cite{Ristani:CVPR2018}. 
Market1501 has total 12,936 images of 751 identities in the training set, and in total 19,732 images of 750 identities; whereas DukeMTMC-reID has total 16,522 images of 702 identities in the training set, and in total 19,989 images of 702 identities.

\myparagraph{Evaluation Metrics}
We use 
two 
popular metrics for person ReID, including Cumulative Match Characteristic (CMC) and mean Average precision (mAP).
For CMC, we only use top-1 to evaluate the performance. 

\subsection{Implementation Details}

Our MGCE-HCL adopts ImageNet to pretrain ResNet50 as the backbone.
Each input is resized to $256\times128$, and is transformed by horizontal flip and random erasing \cite{zhong2020RandomErasing}, whose probabilities are all set to 0.5.
The range of the parameter $d$ in MGCE are set as $[0.4, 0.6]$ with interval $\delta =0.05$.  
The parameter $\tau$ in the loss $\mathcal{L}_{PC}$ in Eq.~\eqref{eq:PC-loss} is set to 0.05 and the momentum parameter $\gamma$ in Eq.~\eqref{eq:memory-update} is set to 0.2. 
During the training, following the protocol in the prior work, we select 16 \textit{pseudo} identities\footnote{We should note that the pseudo identities are obtained from clustering result rather than using the ground-truth labels.} and 4 samples per identity as 
each mini-batch, and train totally 50 epochs. 
In experiments, we utilize the Adam optimizer \cite{Kingma:arXiv2014} to optimize the network with a weight decay rate 
$5\times 10^{-4}$.

\subsection{Ablation Study}
To validate the effectiveness of each component in our proposals, we conduct a set of 
ablation experiments. 

\myparagraph{HCL vs. Cluster-level and Instance-level based Methods}
In HCL, 
we use the negative samples without any clustering structural information. 
To validate 
the effectiveness of our approach, we compare our HCL approach 
with the cluster-level contrastive learning method, which is marked as ``clusterNCE'', and the instance-level contrastive learning method which is represented by MoCo \cite{MoCov1}. 
We conduct a set of experiments with the commonly used best-performing 
parameter $d$ in DBSCAN as 
in prior works \cite{Ge:NIPS20,GE:ICLR20,zhai:ECCV20,Wang:AAAI2021-CAP}. 
Experimental results are reported in Table~\ref{table:CLUB}. We can read from the table that our HCL 
outperforms the instance-level and cluster-level contrastive learning methods in all cases. This is because that 
our hybrid contrastive learning paradigm can 
well grasp the useful information and effectively 
eliminate the impact of negative samples compared to MoCo and clusterNCE. 

\begin{table}
    \caption{HCL vs. clusterNCE and MoCo. The results of MoCo are cited from \cite{Ge:NIPS20}.}
    \centering
    \begin{threeparttable}
    \resizebox{0.6\textwidth}{!}{
    \begin{tabular}{l|c c|c c}
    \hline
    \multirow{2}{*}{Method}&\multicolumn{2}{c|}{Market-1501} & \multicolumn{2}{c}{DukeMTMC-reID} \\
           & mAP  &top-1  &mAP  &top-1\\
    \hline
    \hline
    MoCo\cite{MoCov1} &6.1&12.8&5.6&10.7 \\
    \hline
    clusterNCE ($d=0.4$) &69.2&86.8&59.0&76.2 \\
    \textbf{HCL ($d=0.4$)} &74.6&89.4&61.1&77.2 \\
    \hline
    clusterNCE ($d=0.5$) &73.9&87.9&63.6&79.2 \\
    \textbf{HCL ($d=0.5$)} &\textbf{79.4}&\textbf{91.7}&\underline{66.2}&\underline{81.3} \\
    \hline
    clusterNCE ($d=0.6$) &68.9&85.5&62.5&78.6 \\
    \textbf{HCL ($d=0.6$)} &\underline{77.2}&\underline{90.1}&\textbf{67.4}&\textbf{81.8} \\
    \hline
    \end{tabular}
    }
    \label{table:CLUB}
    \end{threeparttable}
\end{table}

\begin{table}
\caption{Comparison of MGCE-HCL and HCL with different $d$}
\centering
\begin{threeparttable}
\resizebox{0.70\textwidth}{!}{
\begin{tabular}{l|c c | c c}
\hline
    \multirow{2}{*}{Method}&\multicolumn{2}{c|}{Market-1501} & \multicolumn{2}{c}{DukeMTMC-reID} \\
           & mAP  &top-1  &mAP  &top-1\\
\hline
\hline
HCL ($d=0.40$) &74.6&89.4&61.1&77.2 \\
HCL ($d=0.45$) &77.4&90.9&63.3&78.2 \\
HCL ($d=0.50$) &\underline{79.4}&\underline{91.7}&66.2&81.3 \\
HCL ($d=0.55$) &79.0&91.2&67.0&\textbf{82.5} \\
HCL ($d=0.60$) &77.2&90.1&\underline{67.4}&\underline{81.8} \\
\hline
MGCE-HCL ($d \in [0.4,0.6]$)&\textbf{79.6}&\textbf{92.1}&\textbf{67.5}&\textbf{82.5}\\
\hline
\end{tabular}
}
\end{threeparttable}
\label{table:MGCE}
\end{table}

\myparagraph{MGCE-HCL vs. HCL} 
In HCL, we perform DBSCAN with a fixed neighborhood radius parameter $d$ 
to obtain a specific clustering result and thus the pseudo labels; whereas in MGCE, 
we perform DBSCAN for multiple times, each time with a different parameter $d$, to obtain multiple clustering results. 
%
In previous DBSCAN based methods, 
it has been reported that the best-performing results are obtained usually 
when $d$ is taking the value from 0.4 to 0.6. 
This means that performing clustering with $d$ in such an interval will gain the most useful 
clustering information.
To demonstrate 
the effectiveness of our clustering ensemble strategy, we compare our MGCE-HCL with HCL, 
in which both of them use $d$ taking from $0.4$ to $0.6$ in an interval of $0.05$. The experimental results are shown in Table~ \ref{table:MGCE}. 
We can observe that HCL 
performs best when $d$ is set to 0.5 for Market-1501 and 0.55 or 0.6 for DukeMTMC; whereas MGCE-HCL
outperforms all the cases of HCL 
on both Market-1501 and DukeMTMC-reID.

\begin{table}
    \caption{Evaluation on MGCE-HCL with different parameter ranges.}
    \centering
    \begin{threeparttable}
    \resizebox{0.55\textwidth}{!}{
    \begin{tabular}{l|c c|c c}
    \hline
    \multirow{2}{*}{Ensemble Range}&\multicolumn{2}{c|}{Market-1501} & \multicolumn{2}{c}{DukeMTMC-reID} \\
           & mAP  &top-1  &mAP  &top-1\\
    \hline
    \hline
    0.1-0.3 &39.3&63.3&48.8&67.6 \\
    0.2-0.3 &49.5&74.0&51.8&71.1 \\
    \hline
    0.1-0.4 &65.6&84.7&58.9&75.9 \\
    0.2-0.4 &68.2&86.8&58.4&75.5 \\
    0.3-0.4 &73.3&89.1&60.1&77.0 \\
    \hline
    0.1-0.5 &75.8&90.3&63.4&79.7 \\
    0.2-0.5 &77.5&91.3&63.5&79.2 \\
    0.3-0.5 &78.2&91.1&64.6&80.1 \\
    0.4-0.5 &79.3&91.4&64.9&80.7 \\
    
    \hline
    0.1-0.6 &79.0&91.5&67.0&81.7 \\
    0.2-0.6 &79.1&91.0&67.2&81.9 \\
    0.3-0.6 &79.4&91.8&66.8&81.3 \\
    0.4-0.6 &\textbf{79.6}&\textbf{92.1}&\textbf{67.5}&\textbf{82.5} \\
    0.5-0.6 &78.1&91.3&67.3&81.5 \\
    
    \hline
    0.3-0.7 &17.7&35.4&7.8&15.1 \\
    0.4-0.7 &50.0&74.0&46.8&64.7 \\
    \hline
    \end{tabular}
    }
    \label{table:area}
    \end{threeparttable}
\end{table}

\begin{table}
\caption{Evaluation on MGCE-HCL with different interval $\delta$.}
\centering
\begin{threeparttable}
\resizebox{0.45\textwidth}{!}{
\begin{tabular}{l|c c | c c}
\hline
    \multirow{2}{*}{Interval $\delta$}&\multicolumn{2}{c|}{Market-1501} & \multicolumn{2}{c}{DukeMTMC-reID} \\
           & mAP  &top-1  &mAP  &top-1\\
\hline
\hline
0.05 &\textbf{79.6}&\textbf{92.1}&\textbf{67.5}&\textbf{82.5} \\
0.02 &78.8&91.2&67.0&81.8 \\
0.01 &79.5&91.7&67.2&81.5 \\
\hline
\end{tabular}
}
\end{threeparttable}
\label{table:granularity}
\end{table}

\begin{table}
\caption{Evaluation on MGCE-HCL with Different $\gamma$.}
\centering
\begin{threeparttable}
\resizebox{0.45\textwidth}{!}{
\begin{tabular}{l|c c | c c}
\hline
    \multirow{2}{*}{$\gamma$}&\multicolumn{2}{c|}{Market-1501} & \multicolumn{2}{c}{DukeMTMC-reID} \\
           & mAP  &top-1  &mAP  &top-1\\
\hline
\hline
0.1 &78.8&91.1&67.1&81.6 \\
0.2 &\textbf{79.6}&\textbf{92.1}&\textbf{67.5}&\textbf{82.5} \\
0.3 &\textbf{79.6}&92.0&67.0&81.4 \\
0.4 &79.3&91.7&66.0&80.5 \\
0.5 &79.0&91.4&64.3&79.9 \\
\hline
\end{tabular}
}
\end{threeparttable}
\label{table:gamma}
\end{table}



\begin{table}[htbp]
    \caption{Comparison to SOTA methods on Market-1501 and DukeMTMC. 
    }
    \centering
    \begin{threeparttable}
    \resizebox{0.65\textwidth}{!}{
    \begin{tabular}{c|l|c|c c | c c}
    \hline
    \multirow{2}{*}{Type} &\multirow{2}{*}{Method} &\multirow{2}{*}{Reference} &\multicolumn{2}{c|}{Market-1501} &\multicolumn{2}{c}{DukeMTMC-reID} \\
    &                        &                           &mAP &top-1  &mAP & top-1\\
    \hline
    \hline
    \multirow{6}{*}{\shortstack{UDA}} 
    &PTGAN \cite{Wei:CVPR18}     &CVPR’18 &15.7&38.6 &13.5&27.4        \\
    &SPGAN \cite{Deng:CVPR18}    &CVPR’18 &26.7&58.1 &26.4&46.9  \\
    &HHL\cite{Zhong:ECCV18} &ECCV'18 &31.4&62.2&27.2&46.9 \\
    &ECN\cite{Zhong:CVPR19} &CVPR'19 &43.0&75.1&40.4&63.3 \\
    &SSG \cite{Fu:ICCV19}  &ICCV’19    &58.3&80.0&53.4&73.0        \\
    &MMCL\cite{Wang:CVPR20} &CVPR'20 
 &60.4  &84.4 &51.4&72.4 \\
 	&ECN++\cite{zhong:TPAMI2020} &TPAMI'20 &63.8&84.1&54.4&74.0 \\
    &AD-cluster\cite{zhai:CVPR20} &CVPR'20   
 &68.3  &86.7   &54.1&72.6 \\
    &MMT\cite{GE:ICLR20} &ICLR'20        &73.8&89.5&62.3&76.3 \\
    &SpCL\cite{Ge:NIPS20} &NeuIPS'20        &76.7&90.3&68.8&82.9 \\
    \hline
     \multirow{13}{*}{\shortstack{USL}}
      &LOMO\cite{Liao:CVPR15} &CVPR'15
  &8.0&27.2&4.8&12.3 \\
      &BoW\cite{Zheng:ICCV15} &ICCV'15
  &14.8&35.8&8.5&17.1 \\
     &PUL\cite{Fan:TOMM18} &TOMM'18
  &22.8&51.5&22.3&41.1 \\
     &CAMEL\cite{Yu:ICCV17} &ICCV'17
  &26.3&54.4&19.8&40.2 \\
     &BUC\cite{Lin:AAAI19} &AAAI'19
  &30.6&61.0&21.9&40.2 \\
     &SSL\cite{Lin:CVPR20} &CVPR'20
  &37.8&71.7&28.6&52.5 \\
     &HCT\cite{Zeng:CVPR20} &CVPR'20
  &56.4&80.0&50.1&69.6 \\
     &SpCL\cite{Ge:NIPS20} &NeurIPS'20
  &72.6&87.7&65.3&81.2 \\
  &CAP\cite{Wang:AAAI2021-CAP} &AAAI'21
  &\underline{79.2}&\underline{91.4}&67.3&81.1 \\
    \cline{2-7}
   &\textbf{HCL} &This paper&77.2&90.1 & \underline{67.4}& \underline{81.8} \\
   &\textbf{MGCE-HCL}     &This paper   
    &\textbf{79.6}&\textbf{92.1} &\textbf{67.5}&\textbf{82.5} \\

    \hline
    \end{tabular}
    }
    \end{threeparttable}
    \vspace{-10pt}
    \label{table:SOTA}
    \end{table}

\myparagraph{Evaluation on Parameter Range for MGCE-HCL} 
%
To explore the proper range to sample the parameter $d$, we conduct experiments with $d$ sampled in different ranges and 
show the results in Table~\ref{table:area}. 
Since that the parameter $d$ is used to determine the neighborhood, it is not reasonable to set it too large and the same for the upper bound of the parameter range in MGCE. According to the experience, when $d$ is set in the range of $[0.4, 0.6]$, the cluster results might combine positive samples and moderate noises which contain rich and reliable clustering information. To make full use of such clustering information in the range of $[0.4,0.6]$, we set the upper bound of the parameter range as $0.5,0.6,0.7$, respectively, and increase the lower bound of the range from $0.1$ and using an interval $\delta=0.05$ for fair comparison. We also add the experiments with the upper bound of 0.4 to validate the robustness of our MGCE-HCL. Experiments are shown in Table~\ref{table:area}. We can read that when the upper bound is $0.6$, MGCE-HCL 
yields better performance. Especially when the lower bound is set as $0.4$, MGCE-HCL achieves the best performance. This result suggests 
that the range of $[0.4, 0.6]$ for the parameter $d$ to perform DBSCAN contains the richest clustering information and it is consistent with the common practice 
for setting the parameter $d$ in prior works \cite{Ge:NIPS20,GE:ICLR20,zhai:ECCV20,Wang:AAAI2021-CAP}. 
%
%
Moreover, we find that decreasing the lower bound may cause slight drop on the performance. The reason 
is that when the upper bound is fixed, decreasing the lower bound will decrease the priority of samples from clusters of larger size, which may contain more useful information.
It is worth to note that MGCE-HCL is insensitive to the lower bound of ensemble range and not that sensitive to the upper bound of the range when the upper bound is not over-large. This hints that we can obtain reasonably good performance with a relatively larger 
range for clustering ensemble 
and an appropriate upper bound even if we do not know the exact best parameter $d$. 
However, as shown in Table~\ref{table:area}, the results might sharply drop when we set the upper bound up to 0.7.
It is because that the clustering results will be too noisy when using an over-large parameter $d$.

\myparagraph{Evaluation on $\delta$ in Cluster Ensemble}
To evaluate the effect of the interval in the ensemble range, 
we fix the parameter range to pick 
$d$ as $[0.4, 0.6]$ and 
change the sampling interval $\delta$ to $\{ 0.05, 0.02, 0.01\}$, individually. Experimental results are shown in Table~\ref{table:granularity}. Using a smaller $\delta$ leads to a larger $T$, \ie, the times of running DBSCAN. The results show that MGCE-HCL is also insensitive to the interval $\delta$. 


\myparagraph{Evaluation on momentum factor $\gamma$}
As shown in Eq.~\ref{eq:memory-update}, parameter $\gamma$ is the momentum factor to update memory bank. 
In our method, memory bank is used to store relatively static features (\ie smoothed features), 
rather than using the 
features directly extracted from the output of the backbone. Therefore, the momentum factor $\gamma$ should not be too large. 
To evaluate the effect of using different parameter $\gamma$, we conduct a set of experiments to compare the performance with different $\gamma$ in Table~\ref{table:gamma}. 
The results show that $\gamma=0.2$ achieves the best performance and the performance gradually drops when $\gamma$ is larger than 0.2.

\subsection{Comparison to State-of-the-art Methods}

Finally, we compare the performance of our proposed MGCE-HCL method to the state-of-the-art methods on Market-1501 and DukeMTMC-reID. The experimental 
results are shown in Table ~\ref{table:SOTA}. 

\myparagraph{Compared to USL-based methods} We compare the most unsupervised works recent years, including BoW \cite{Zheng:ICCV15}, LOMO \cite{Liao:CVPR15}, PUL \cite{Fan:TOMM18}, CAMEL \cite{Yu:ICCV17}, BUC \cite{Lin:AAAI19}, SSL \cite{Lin:CVPR20}, HCT \cite{Zeng:CVPR20}, SpCL \cite{Ge:NIPS20} and CAP \cite{Wang:AAAI2021-CAP}. 
In the USL-based methods, only the unlabeled data is used. 
In most recent works, \eg, BUC, HCT, SpCL and CAP, they are 
based on a clustering method (\eg, $k$-means, DBSCAN, or hierarchical clustering) to yield the pseudo labels.
Among them, SpCL and CAP also use cluster-based contrastive learning method. SpCL learns with a self-paced strategy and CAP introduces the camera information to boost training. Compared to CAP, our 
HCL obtains the comparable results on Market-1501 and obtains 0.7\% top-1 and 0.1\% mAP performance gain on DukeMTMC-reID, and 
our MGCE-HCL obtains 0.7\% top-1 and 0.4\% mAP performance gain on Market-1501, and obtains 1.4\% top-1 gain and 0.2\% mAP gain on DukeMTMC-reID, respectively.

\myparagraph{Compared to UDA-based methods} We also list the results for UDA based methods 
at the upper part in Table~\ref{table:SOTA}. The UDA-based methods exploit the information from source domain to improve the performance of unlabeled target domain. 
In the UDA part, the column of Market-1501 shows the results where model is transferred from DukeMTMC-reID to Market-1501, and vice versa for the column of DukeMTMC-reID.
Note that our methods without any label annotation can outperform SpCL on Market-1501 and are on par with SpCL on DukeMTMC-reID.

\section{Conclusions}


We have proposed a hybrid contrastive learning (HCL) paradigm for unsupervised person ReID, in which the cluster structure of positive samples are reserved but the clusters of the negative samples are ignored. 
Moreover, we have presented a multi-granularity cluster ensemble (MGCE) approach to weight the positive samples in different granularity with a priority, and developed a priority-weighted hybrid contrastive loss for training, by which the noises especially from larger granularity clusters can be reduced to some extent.
We conducted extensive experiments on two benchmark datasets and the results shown that our HCL paradigm notably outperforms the instance-level contrastive learning paradigm and cluster-level contrastive learning paradigm, and our MGCE-HCL approach achieves the better performance compared to state-of-the-art methods.



\section*{Acknowledgment}
This work is supported by the National Natural Science Foundation of China under Grant 61876022.

%
%
%
%
{
\bibliographystyle{splncs04}
\bibliography{main}

\begin{thebibliography}{10}
\providecommand{\url}[1]{\texttt{#1}}
\providecommand{\urlprefix}{URL }
\providecommand{\doi}[1]{https://doi.org/#1}

\bibitem{simclr}
Chen, T., Kornblith, S., Norouzi, M., Hinton, G.: A simple framework for
  contrastive learning of visual representations. In: International conference
  on machine learning. pp. 1597--1607. PMLR (2020)

\bibitem{Deng:CVPR18}
Deng, W., Zheng, L., Ye, Q., Kang, G., Yang, Y., Jiao, J.: Image-image domain
  adaptation with preserved self-similarity and domain-dissimilarity for person
  re-identification. In: IEEE Conference on Computer Vision and Pattern
  Recognition. pp. 994--1003 (2018)

\bibitem{EsterDBSCAN:AAAI96}
Ester, M., Kriegel, H.P., Sander, J., Xu, X.: A density-based algorithm for
  discovering clusters in large spatial databases with noise. In: Proceedings
  of the Second International Conference on Knowledge Discovery and Data
  Mining. p. 226–231 (1996)

\bibitem{Fan:TOMM18}
Fan, H., Zheng, L., Yan, C., Yang, Y.: Unsupervised person re-identification:
  Clustering and fine-tuning. ACM Transactions on Multimedia Computing,
  Communications, and Applications  \textbf{14}(4), ~83 (2018)

\bibitem{Fu:ICCV19}
Fu, Y., Wei, Y., Wang, G., Zhou, Y., Shi, H., Huang, T.S.: Self-similarity
  grouping: A simple unsupervised cross domain adaptation approach for person
  re-identification. In: Proceedings of the IEEE/CVF International Conference
  on Computer Vision. pp. 6112--6121 (2019)

\bibitem{GE:ICLR20}
Ge, Y., Chen, D., Li, H.: Mutual mean-teaching: Pseudo label refinery for
  unsupervised domain adaptation on person re-identification. In: International
  Conference on Learning Representations (2020)

\bibitem{Ge:NIPS20}
Ge, Y., Zhu, F., Chen, D., Zhao, R., Li, H.: Self-paced contrastive learning
  with hybrid memory for domain adaptive object re-id. In: Advances in Neural
  Information Processing Systems (2020)

\bibitem{grill2020:BYOL}
Grill, J.B., Strub, F., Altch{\'e}, F., Tallec, C., Richemond, P.H.,
  Buchatskaya, E., Doersch, C., Pires, B.A., Guo, Z.D., Azar, M.G., et~al.:
  Bootstrap your own latent: A new approach to self-supervised learning. arXiv
  preprint arXiv:2006.07733  (2020)

\bibitem{2010NCE}
Gutmann, M., Hyvärinen, A.: Noise-contrastive estimation: A new estimation
  principle for unnormalized statistical models. Journal of Machine Learning
  Research  \textbf{9},  297--304 (2010)

\bibitem{MoCov1}
He, K., Fan, H., Wu, Y., Xie, S., Girshick, R.: Momentum contrast for
  unsupervised visual representation learning. In: Proceedings of the IEEE/CVF
  Conference on Computer Vision and Pattern Recognition. pp. 9729--9738 (2020)

\bibitem{He:CVPR2016:resnet}
He, K., Zhang, X., Ren, S., Sun, J.: Deep residual learning for image
  recognition. In: Proceedings of the IEEE conference on computer vision and
  pattern recognition. pp. 770--778 (2016)

\bibitem{Kingma:arXiv2014}
Kingma, D.P., Ba, J.: Adam: A method for stochastic optimization. arXiv
  preprint arXiv:1412.6980  (2014)

\bibitem{Liao:CVPR15}
Liao, S., Hu, Y., Zhu, X., Li, S.Z.: Person re-identification by local maximal
  occurrence representation and metric learning. In: IEEE Conference on
  Computer Vision and Pattern Recognition. pp. 2197--2206 (2015)

\bibitem{Lin:AAAI19}
Lin, Y., Dong, X., Zheng, L., Yan, Y., Yang, Y.: A bottom-up clustering
  approach to unsupervised person re-identification. In: The Association for
  the Advancement of Artificial Intelligence. vol.~33, pp. 8738--8745 (2019)

\bibitem{Lin:CVPR20}
Lin, Y., Xie, L., Wu, Y., Yan, C., Tian, Q.: Unsupervised person
  re-identification via softened similarity learning. In: IEEE Conference on
  Computer Vision and Pattern Recognition (2020)

\bibitem{Oord2018:CPC}
Oord, A.v.d., Li, Y., Vinyals, O.: Representation learning with contrastive
  predictive coding. arXiv preprint arXiv:1807.03748  (2018)

\bibitem{Ristani:CVPR2018}
Ristani, E., Tomasi, C.: Features for multi-target multi-camera tracking and
  re-identification. In: Proceedings of the IEEE conference on computer vision
  and pattern recognition. pp. 6036--6046 (2018)

\bibitem{Wang:CVPR20}
Wang, D., Zhang, S.: Unsupervised person re-identification via multi-label
  classification. In: Proceedings of the IEEE/CVF Conference on Computer Vision
  and Pattern Recognition. pp. 10981--10990 (2020)

\bibitem{wang2018:mgn}
Wang, G., Yuan, Y., Chen, X., Li, J., Zhou, X.: Learning discriminative
  features with multiple granularities for person re-identification. In:
  Proceedings of the 26th ACM international conference on Multimedia. pp.
  274--282 (2018)

\bibitem{Wang:AAAI2021-CAP}
Wang, M., Lai, B., Huang, J., Gong, X., Hua, X.S.: Camera-aware proxies for
  unsupervised person re-identification. In: Proceedings of the AAAI Conference
  on Artificial Intelligence (AAAI) (2021)

\bibitem{Wei:CVPR18}
Wei, L., Zhang, S., Gao, W., Tian, Q.: Person transfer gan to bridge domain gap
  for person re-identification. In: IEEE Conference on Computer Vision and
  Pattern Recognition. pp. 79--88 (2018)

\bibitem{wu2018unsupervised}
Wu, Z., Xiong, Y., Yu, S.X., Lin, D.: Unsupervised feature learning via
  non-parametric instance discrimination. In: Proceedings of the IEEE
  Conference on Computer Vision and Pattern Recognition. pp. 3733--3742 (2018)

\bibitem{Yu:ICCV17}
Yu, H.X., Wu, A., Zheng, W.S.: Cross-view asymmetric metric learning for
  unsupervised person re-identification. In: IEEE International Conference on
  Computer Vision. pp. 994--1002 (2017)

\bibitem{yu2019:mar}
Yu, H.X., Zheng, W.S., Wu, A., Guo, X., Gong, S., Lai, J.H.: Unsupervised
  person re-identification by soft multilabel learning. In: Proceedings of the
  IEEE/CVF Conference on Computer Vision and Pattern Recognition. pp.
  2148--2157 (2019)

\bibitem{Zeng:CVPR20}
Zeng, K., Ning, M., Wang, Y., Guo, Y.: Hierarchical clustering with hard-batch
  triplet loss for person re-identification. In: IEEE Conference on Computer
  Vision and Pattern Recognition. pp. 13657--13665 (2020)

\bibitem{zhai:CVPR20}
Zhai, Y., Lu, S., Ye, Q., Shan, X., Chen, J., Ji, R., Tian, Y.: Ad-cluster:
  Augmented discriminative clustering for domain adaptive person
  re-identification. In: Proceedings of the IEEE/CVF Conference on Computer
  Vision and Pattern Recognition. pp. 9021--9030 (2020)

\bibitem{zhai:ECCV20}
Zhai, Y., Ye, Q., Lu, S., Jia, M., Ji, R., Tian, Y.: Multiple expert
  brainstorming for domain adaptive person re-identification. In: Computer
  Vision--ECCV 2020: 16th European Conference, Glasgow, UK, August 23--28,
  2020, Proceedings, Part VII 16. pp. 594--611. Springer (2020)

\bibitem{Zheng:ICCV15}
Zheng, L., Shen, L., Tian, L., Wang, S., Wang, J., Tian, Q.: Scalable person
  re-identification: A benchmark. In: IEEE International Conference on Computer
  Vision. pp. 1116--1124 (2015)

\bibitem{zhong2020RandomErasing}
Zhong, Z., Zheng, L., Kang, G., Li, S., Yang, Y.: Random erasing data
  augmentation. In: Proceedings of the AAAI Conference on Artificial
  Intelligence. vol.~34, pp. 13001--13008 (2020)

\bibitem{Zhong:ECCV18}
Zhong, Z., Zheng, L., Li, S., Yang, Y.: Generalizing a person retrieval model
  hetero-and homogeneously. In: European Conference on Computer Vision. pp.
  172--188 (2018)

\bibitem{Zhong:CVPR19}
Zhong, Z., Zheng, L., Luo, Z., Li, S., Yang, Y.: Invariance matters: Exemplar
  memory for domain adaptive person re-identification. In: Proceedings of the
  IEEE/CVF Conference on Computer Vision and Pattern Recognition. pp. 598--607
  (2019)

\bibitem{zhong:TPAMI2020}
Zhong, Z., Zheng, L., Luo, Z., Li, S., Yang, Y.: Invariance matters: Exemplar
  memory for domain adaptive person re-identification pp. 598--607 (2019)

\end{thebibliography}
}

\end{document}